\documentclass[twoside,11pt]{article}

\usepackage{paralist,amsmath, amssymb}
\usepackage{booktabs}
\usepackage{multirow}
\usepackage{algorithm,algorithmic}
\usepackage{jmlr2e}

\usepackage[colorlinks,
            linkcolor=blue,
            citecolor=blue,
            urlcolor=magenta,
            linktocpage,
            plainpages=false]{hyperref}

\makeatletter
\newcounter{ALC@tempcntr}
\makeatother

\makeatletter
\AtBeginDocument{\Hy@breaklinkstrue}
\makeatother

\newtheorem{thm}{Theorem}

\newtheorem{ass}{Assumption}

\def \D {\mathcal{D}}

\def \u {\mathbf{u}}
\def \H {\mathcal{H}}
\def \w {\mathbf{w}}
\def \R {\mathbb{R}}

\def \N {\mathcal{N}}
\def \A {\mathcal{A}}

\def \v {\mathbf{v}}

\def \I {\mathcal{I}}

\def \C {\mathcal{C}}

\def \N {\mathbb{N}}

\DeclareMathOperator*{\Reg}{R}
\DeclareMathOperator*{\WAReg}{A-R}
\DeclareMathOperator*{\SAReg}{SA-R}
\DeclareMathOperator*{\DReg}{D-R}

\DeclareMathOperator*{\argmin}{argmin}

\begin{document}

\title{Minimizing Dynamic Regret and Adaptive Regret Simultaneously}

\author{\name Lijun Zhang \email zhanglj@lamda.nju.edu.cn\\
\name Shiyin Lu \email lusy@lamda.nju.edu.cn\\
       \addr National Key Laboratory for Novel Software Technology\\
       Nanjing University, Nanjing 210023, China \\
\name Tianbao Yang \email tianbao-yang@uiowa.edu\\
       \addr Department of Computer Science, The University of Iowa\\
       Iowa City, IA 52242, USA
}

\editor{}

\maketitle

\begin{abstract}%
Regret minimization is treated as the golden rule in the traditional study of online learning. However, regret minimization algorithms tend to converge to the static optimum, thus being suboptimal for changing environments. To address this limitation, new performance measures, including dynamic regret and adaptive regret have been proposed to guide the design of online algorithms. The former one aims to minimize the global regret with respect to a sequence of changing comparators, and the latter one attempts to minimize every local regret with respect to a fixed comparator. Existing algorithms for dynamic regret and adaptive regret are developed independently, and only target one performance measure. In this paper, we bridge this gap by proposing novel online algorithms that are able to minimize the dynamic regret and adaptive regret \emph{simultaneously}. In fact, our theoretical guarantee is even stronger in the sense that one algorithm is able to minimize the dynamic regret over \emph{any} interval.
\end{abstract}


\section{Introduction}
Online convex optimization (OCO) is a powerful framework for sequential decision making and has found a variety of applications \citep{Intro:Online:Convex}. The protocol of OCO can be viewed as a repeated game between a learner and an adversary: In each round $t=1,2,\ldots,T$, the learner selects an action $\w_t$ from a convex feasible set $\Omega$, and at the same time the adversary chooses a convex loss function $f_t(\cdot): \Omega \mapsto \R$. Then, the function is revealed to the learner who incurs an instantaneous loss $f_t(\w_t)$. The goal of the learner is to minimize the regret:
\begin{equation} \label{eqn:static-regret}
  \Reg(T) = \sum_{t=1}^T f_t(\w_t)  - \min_{\w \in \Omega} \sum_{t=1}^T f_t(\w)
\end{equation}
which compares the cumulative loss of the learner to that of the best fixed action in hindsight, and is typically referred to as \emph{static} regret since the comparator is time-invariant.

Over the past decades, static regret has been extensively studied and algorithms with minimax optimal regret bounds have been developed \citep{zinkevich-2003-online,ML:Hazan:2007,ICML_Pegasos,NIPS2007_3319,NIPS2010_Smooth,Online:suvery}. However, the metric of static regret is only meaningful for stationary environments, and low static regret does not necessarily imply a good performance in changing environments since the time-invariant comparator in (\ref{eqn:static-regret}) may behave badly. To address this limitation, recent studies have introduced more stringent performance metrics, including dynamic regret and strongly adaptive regret, to measure the learner's performance.

The dynamic regret is defined as the difference between the cumulative loss of the learner and that of a sequence of comparators $\u_1, \ldots, \u_T \in \Omega$ \citep{zinkevich-2003-online}:
\begin{equation} \label{eqn:dynamic:1}
\DReg(\u_1,\ldots,\u_T) = \sum_{t=1}^T f_t(\w_t)  - \sum_{t=1}^T f_t(\u_t).
\end{equation}
While it is well-known that sublinear dynamic regret is unattainable in the worst case, one can bound the dynamic regret in terms of some regularities of the comparator sequence. A remarkable example is given by \citet{zinkevich-2003-online}, who introduces the notion of path-length  defined in (\ref{eqn:path-length}) to measure the temporal variability of the comparator sequence, and derives an $O(\sqrt{T}(1+P_T))$ dynamic regret bound, where $P_T$ is the path-length. Very recently, \citet{Adaptive:Dynamic:Regret:NIPS} improve this result to be optimal by establishing an $O(\sqrt{T(1+P_T)})$ upper bound as well as a matching lower bound.

The strongly adaptive regret evaluates the learner's performance on each time interval of length $\tau$, and is defined as the maximum static regret over these intervals \citep{Adaptive:ICML:15}:
\begin{equation} \label{eqn:strong:adaptive}
\begin{split}
\SAReg(T,\tau) = \max_{[s, s+\tau -1] \subseteq [T]} \left(\sum_{t=s}^{s+\tau -1}  f_t(\w_t)  - \min_{\w \in \Omega}   \sum_{t=s}^{s+\tau -1}  f_t(\w)  \right).
\end{split}
\end{equation}
In the above definition, since the benchmark action $\w$ that minimizes the cumulative loss over interval $[s, s+\tau -1]$ typically varies with $s$, the learner is essentially competing with changing comparators. The state-of-the-art strongly adaptive regret bound is $O(\sqrt{\tau \log{T}})$ \citep{Improved:Strongly:Adaptive} which matches the minimax static regret over a fixed interval up to a logarithmic factor \citep{Minimax:Online}.

Dynamic regret handles changing environments from a global prospective, as it measures the performance over the whole interval but allows the comparator changes over time. By contrast, adaptive regret takes a local perspective, since it focuses on short intervals with a fixed comparator but allows the interval changes over time. Although \citet{Adaptive:Dynamic:Regret:NIPS} demonstrate that it is possible to derive dynamic regret from adaptive regret, their dynamic regret bound only takes a special form, and thus it does not mean adaptive regret is more fundamental than dynamic regret. Since  dynamic regret and adaptive regret reflect different perspectives and are complementary to each other, it is appealing to ask whether we can minimize these two metrics simultaneously. Unfortunately, existing algorithms for minimizing dynamic regret and adaptive regret are developed independently and differ significantly.

In this paper, we propose novel algorithms that minimize the dynamic regret and adaptive regret simultaneously. Our methods follow the framework of ``prediction with expert advice'' \citep{bianchi-2006-prediction} and share a similar hierarchical structure: a series of expert algorithms configured with different lifetimes running in parallel, and an expert-tracking algorithm that combines the actions of all active experts. Specifically, the first method uses the simple online gradient descent (OGD) as the expert algorithm but manages the lifetime of experts through specifically designed intervals. On the contrary, the second method utilizes standard techniques to activate and deactivate experts, at the cost of a more complicated expert algorithm. Theoretical analysis shows that both methods attain the state-of-the-art $O(\sqrt{\tau \log T})$ adaptive regret, and they achieve $O( \sqrt{T (1+ P_T )\log T })$ and $O( \sqrt{T (\log T+ P_T ) })$ dynamic regrets, respectively. Furthermore, the second method enjoys an even stronger theoretical guarantee: it can minimize the dynamic regret over any interval.
\section{Related Work}
In this section, we briefly review related work in dynamic regret and adaptive regret for OCO.
\subsection{Dynamic Regret}
Dynamic regret is first introduced  by \citet{zinkevich-2003-online}, who proposes to use the \emph{path-length}
\begin{equation} \label{eqn:path-length}
   P_T=\sum_{t=1}^T \|\u_{t+1} - \u_{t}\|_2
\end{equation}
to measure the performance. Specifically, \citet{zinkevich-2003-online} demonstrates  that OGD with a constant step size attains a dynamic regret of  $O(\sqrt{T} (1+P_T))$ for \emph{any} sequence $\u_1,\ldots,\u_T$. This upper bound is adaptive in the sense that it automatically becomes tighter when the comparators change slowly. Another regularity of the comparator sequence  is defined as
\[
P'_T=\sum_{t=1}^T \|\u_{t+1} - \Phi_{t} (\u_{t})\|_2
\]
where $\Phi_t (\cdot)$ is a dynamic model that predicts a reference point for the $t$-th round.  \citet{Dynamic:ICML:13} develope a novel algorithm named dynamic mirror descent and prove a dynamic regret of $O(\sqrt{T} (1+P'_T))$.  An $\Omega(\sqrt{T(1+P_T)})$ lower bound of dynamic regret is established by \citet{Adaptive:Dynamic:Regret:NIPS}, which indicates the results of \citet{zinkevich-2003-online} and \citet{Dynamic:ICML:13} are far away from the optimum.  To address this limitation,   \citet{Adaptive:Dynamic:Regret:NIPS} develop an optimal algorithm, namely adaptive learning for dynamic environment (Ader), which attains an $O(\sqrt{T(1+P_T)})$ bound in the general case, and an $O(\sqrt{T(1+P_T')})$  bound when a sequence of dynamical models is available.

Deviating from the definition in (\ref{eqn:dynamic:1}), most studies on dynamic regret only consider a restricted form, defined with respect to a sequence of  minimizers of the loss functions due to its greater mathematical tractability \citep{Dynamic:AISTATS:15,Non-Stationary,Dynamic:2016,Dynamic:Strongly,Dynamic:Regret:Squared}:
\begin{equation} \label{eqn:dynamic:2}
\begin{split}
\DReg(\w_1^*, \ldots, \w_T^*) =\sum_{t=1}^T f_t(\w_t)  - \sum_{t=1}^T   f_t(\w_t^*)=\sum_{t=1}^T f_t(\w_t)  -\sum_{t=1}^T \min_{\w \in \Omega} f_t(\w)
\end{split}
\end{equation}
where $\w_t^* \in \argmin_{\w \in \Omega} f_t(\w)$ is a minimizer of $f_t(\cdot)$ over domain $\Omega$. Although one can show that $\DReg(\w_1^*, \ldots, \w_T^*) \geq \DReg(\u_1,\ldots,\u_T)$, it does not imply the former one is stronger since an upper bound for $\DReg(\w_1^*, \ldots, \w_T^*)$ could be very loose for $\DReg(\u_1,\ldots,\u_T)$. In fact, the definition in  (\ref{eqn:dynamic:1}) is more general since it holds for any sequence of comparators, and thus includes the static regret in (\ref{eqn:static-regret}) and the restricted dynamic regret in (\ref{eqn:dynamic:2}) as special cases.

Let $P_T^*$ be the path-length of the minimizer sequence $\w_1^*,\ldots,\w_T^*$. When the loss functions are strongly convex and smooth, \citet{Dynamic:Strongly} show that the restricted dynamic regret of OGD is $O(P_T^*)$. This rate is also attainable for convex and smooth functions  under the condition that the minimizers lie in the interior of $\Omega$ \citep{Dynamic:2016}.
\citet{Dynamic:Regret:Squared} introduce the squared path-length:
\[
S_T^* = \sum_{t=1}^T \| \w_{t+1}^* -\w_t^* \|_2^2
\]
which could be much smaller than $P_T^*$ in the case that  minimizers move slowly. They demonstrate that the restricted dynamic regret bound for strongly convex functions could be improved to $O(\min(P_T^*, S_T^*))$.

Instead of measuring the complexity of the comparator sequence, \citet{Non-Stationary} propose to evaluate the movement of the loss functions as follows:
\begin{equation} \label{eqn:function-variation}
   F_T = \sum \limits_{t=1}^T \sup \limits_{\w \in \Omega} \vert f_{t+1}(\w) - f_t(\w) \vert.
\end{equation}
\citet{Non-Stationary} show that a restarted OGD algorithm equipped with a prior knowledge of an upper bound $V_T \geq F_T$ achieves $O(V_T^{1/3} T^{2/3})$ and $O(\sqrt{V_T T})$ dynamic regret for convex functions and strongly convex functions, respectively. However, these bounds depend on the predetermined $V_T$ rather than the actual $F_T$, and thus are not adaptive.

\subsection{Adaptive Regret}
In their seminal work, \citet{Adaptive:Hazan} define adaptive regret as
\begin{equation} \label{eqn:adaptive:regret}
 \WAReg(T)=\max_{[s, q] \subseteq [T]} \left(\sum_{t=s}^{q} f_t(\w_t) - \min_{\w \in \Omega} \sum_{t=s}^{q} f_t(\w) \right)
\end{equation}
which is the maximum  regret over any contiguous interval. They develop a novel algorithm named as follow the leading history (FLH), which runs an instance of low-regret algorithm in each round as an expert, and then combines them with an expert-tracking method. To improve the efficiency, \citet{Adaptive:Hazan} deploy a data-streaming technique to prune the set of experts, and as a result only $O(\log t)$ experts are stored at round $t$. The efficient version of FLH  attains $O(d \log^2 T)$  and $O(\sqrt{T \log^3 T})$ adaptive regrets for exponentially concave functions and convex functions, respectively \citep{Hazan:2009:ELA}.

However, the adaptive regret in (\ref{eqn:adaptive:regret}) does not respect short intervals well. For example, the $O(\sqrt{T \log^3 T})$ adaptive regret of convex functions is vacuous for intervals of size $O(\sqrt{T})$. To avoid this limitation, \citet{Adaptive:ICML:15} propose the strongly adaptive regret in (\ref{eqn:strong:adaptive}), which emphasizes the dependence on the interval length. The strongly adaptive algorithm of \citet{Adaptive:ICML:15} shares a similar structure to that of FLH \citep{Adaptive:Hazan}, but with the following differences.
\begin{compactenum}[(i)]
  \item \citet{Adaptive:ICML:15} construct a set of geometric covering (GC) intervals, and run an instance of low-regret algorithm for each interval as an expert.
  \item A new meta-algorithm named as strongly adaptive online learner (SOAL) is used to combine  experts.
\end{compactenum}
The GC intervals are defined as
\begin{equation} \label{eqn:GC}
 \I= \bigcup_{k \in \N \cup \{0\}} \I_k
 \end{equation}
 where for all $k \in \N \cup \{0\}$, $\I_k=\left\{ [ i \cdot 2^k, (i+1) \cdot 2^k -1]: i \in \N\right\}$. 
For convex functions, \citet{Adaptive:ICML:15} establish an $O( \sqrt{\tau} \log T )$ strongly adaptive regret. In a subsequent work,  \citet{Improved:Strongly:Adaptive} design a new meta-algorithm named as sleeping coin betting (CB), and improve the strongly adaptive regret to $O( \sqrt{\tau \log T} )$. The adaptive regret of convex and smooth functions are studied by \citet{jun2017} and \citet{Adaptive:Regret:Smooth:ICML}.

\subsection{The Relationship between Dynamic Regret and Adaptive Regret}
In the setting of prediction with expert advice (PEA), dynamic regret is usually referred to as tracking regret or shifting regret \citep{LITTLESTONE1994212,Herbster1998,Track_Large_Expert}. In this case, it has been proved that the tracking regret can be derived from the adaptive regret \citep{Adamskiy2012,Fixed:Share:NIPS12,Adaptive:ICML:15}. In particular, Theorem 4 of \citet{pmlr-v40-Luo15} indicates that it is possible to bound the dynamic regret by the adaptive regret and the following variation:
\[
V_T=  \sum_{t=1}^T \sum_{i=1}^N [u_{t+1,i} - u_{t,i}]_+
\]
where $N$ is the number of experts,  $u_{t,i}$ is the $i$-th component of $\u_t$, and $[x]_+=\max(0,x)$.  Thus, for PEA, it is commonly believed that adaptive regret is more fundamental.

In the setting of OCO, to the best of our knowledge, there is only one work \citep{Dynamic:Regret:Adaptive} that has investigated the relationship between dynamic regret and adaptive regret. Let $I_1 = [s_1, q_1], \ldots, I_k = [s_k, q_k]$ be a partition of $[1, T]$ and for each interval $I_i$, define the local variation of functions as
\[
   F_T(i) = \sum_{t={s_i}}^{q_i-1} \sup \limits_{\w \in \Omega} \vert f_{t+1}(\w) - f_t(\w) \vert.
\]
\citet{Dynamic:Regret:Adaptive} prove that the restricted dynamic regret can be upper bounded in terms of the strongly adaptive regret and $F_T(i)$ as follows:
\[
\begin{split}
\DReg(\w_1^*, \ldots, \w_T^*)  \leq  \min_{I_1, \ldots, I_k} \sum_{i=1}^k \big( \SAReg(T,\vert I_i \vert) + 2 \vert I_i \vert \cdot F_T(i) \big) .
\end{split}
\]
As can be seen, this result is only applicable to the restricted dynamic regret instead of the general dynamic regret considered in this paper.

Following the analysis of \citet{Dynamic:Regret:Adaptive}, we have tried to upper bound the dynamic regret by the strongly adaptive regret and the path-length. 
\begin{thm} \label{thm:rel} Assume all the online functions are $G$-lipschitz continuous, we have
\begin{equation}\label{eqn:dynamic:adap}
\begin{split}
\DReg(\u_1,\ldots,\u_T)\leq \min_{I_1, \ldots, I_k} \sum_{i=1}^k \big(\SAReg(T,|I_i|) + G | I_i | \cdot P_T(i)\big)
\end{split}
\end{equation}
where $P_T(i)=\sum_{t={s_i}}^{q_i-1} \|\u_{t+1} - \u_{t}\|_2$. Combining with the adaptive regret of convex functions \citep{Improved:Strongly:Adaptive}, we obtain the following dynamic regret for convex functions
\begin{equation}\label{eqn:dynamic:by:adap}
\begin{split}
\DReg(\u_1,\ldots,\u_T) =O \left( \max\left\{  \sqrt{T \log T} , T^{2/3} P_T^{1/3} \log^{1/3} T  \right\} \right).
\end{split}
\end{equation}
\end{thm}
The above theorem shows that although the strongly adaptive regret can be used to control the dynamic regret, it may not be able to give the optimal result, since the regret bound in (\ref{eqn:dynamic:by:adap}) is much worse than the $O(\sqrt{T(1+P_T)})$ bound of \citet{Adaptive:Dynamic:Regret:NIPS}. Thus, in the setting of OCO, which one of dynamic regret and adaptive regret is more fundamental remains an open problem. Note that our algorithms are able to minimize the dynamic regret and adaptive regret simultaneously. So, no matter which performance measure is stronger, they are always meaningful.

\section{Our Methods}
In this section, we present our online algorithms that are able to minimize the dynamic regret and adaptive regret simultaneously. The first method uses a two-layer structure, but with specifically designed components. In contrast, the second method has a three-layer structure, but with standard techniques that are easy to comprehend.

\begin{ass}\label{ass:1} The gradients of all functions are bounded by $G$, i.e.,
\begin{equation}\label{eqn:grad}
\max_{\w \in \Omega}\|\nabla f_t(\w)\|_2 \leq G, \ \forall t \in [T].
\end{equation}
\end{ass}
\begin{ass}\label{ass:2} The domain $\Omega$ contains the origin $\mathbf{0}$, and its diameter is bounded by $D$, i.e.,
\begin{equation}\label{eqn:domain}
\max_{\w, \w' \in \Omega} \|\w -\w'\|_2 \leq D.
\end{equation}
\end{ass}
\begin{ass}\label{ass:6} The value of each function belongs to $[0,1]$, i.e.,
\[
0 \leq f_t(\w) \leq 1,  \ \forall \w\in \Omega, t \in [T].
\]
\end{ass}
As long as the loss functions are bounded, they can always be scaled and restricted to $[0, 1]$.
\subsection{The First Method}
\begin{algorithm}[tb]
   \caption{Online Gradient Descent (OGD)}
   \label{alg:ogd}
\begin{algorithmic}[1]
   \STATE {\bfseries Input:} Initial point $\w_1$, and step size $\eta$
   \FOR{$t=1$ {\bfseries to} $T$}
   \STATE Submit $\w_t$, and then receive $f_t(\cdot)$
   \STATE Suffer a loss $f_t(\w_t)$ and update as
   \[
   \w_{t+1} = \Pi_{\Omega}\big[\w_t - \eta \nabla f_t(\w_t)\big]
   \]
   \ENDFOR
\end{algorithmic}
\end{algorithm}
Our first method follows the framework of adaptive algorithms for convex functions \citep{Adaptive:ICML:15,Improved:Strongly:Adaptive}. On one hand, the proposed method inherits their ability to minimize the adaptive regret. On the other hand, our method contains new features so that the dynamic regret can also be minimized.

We take the classical online gradient descent (OGD) as the expert algorithm, and present the procedure in Algorithm~\ref{alg:ogd}. After receiving the loss function $f_t(\cdot)$, OGD performs gradient descent to update the current solution $\w_t$:
\[
\w_{t+1} = \Pi_{\Omega}\big[\w_t - \eta \nabla f_t(\w_t)\big]
\]
where $\Pi_{\Omega}[\cdot]$ denotes the projection onto the nearest point in $\Omega$, and $\eta>0$ is the step size. The following static regret bound of OGD is well-known \citep{zinkevich-2003-online}.
\begin{thm} \label{thm:regret:OGD} Under Assumptions~\ref{ass:1} and \ref{ass:2}, we have
\[
\sum_{t=1}^T f_t(\w_t) - \min_{\w \in \Omega}\sum_{t=1}^T f_t(\w) \leq \frac{D^2}{2 \eta} + \frac{\eta T G^2}{2}=DG \sqrt{T}
\]
where the step size is set as $\eta=D/(G \sqrt{T})$.
\end{thm}

\begin{figure*}
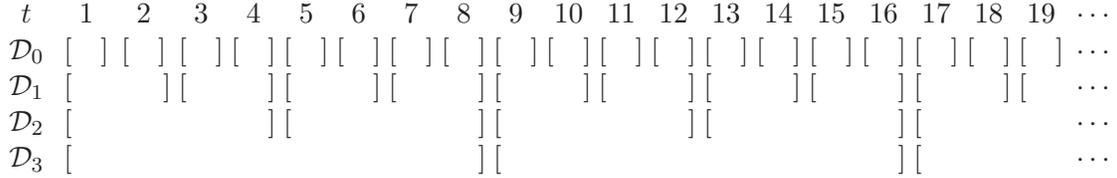

\centering
\begin{tabular}{@{}c@{\hspace{1ex}}*{19}{@{\hspace{0.6ex}}c}@{\hspace{1ex}}c@{}}
$t$ & 1 & 2 & 3 & 4 & 5 & 6 & 7 & 8 & 9 & 10 &11 &12 & 13 & 14 &15 & 16 & 17 & 18 &19  &$\cdots$ \\
 $\D_0$ & [\quad ] & [\quad ] &  [\quad ] & [\quad ] & [\quad ] & [\quad ] & [\quad ] & [\quad ] & [\quad ] & [\quad ] & [\quad ] &[\quad ] &[\quad ] & [\quad ] & [\quad ] & [\quad ]& [\quad ]&[\quad ] & [\quad ]&  $\cdots$   \\
 $\D_1$  & [\quad  \phantom{]}& \phantom{[} \quad ] & [\quad \phantom{]} & \phantom{[}\quad ] & [\quad \phantom{]} & \phantom{[}\quad ] & [\quad \phantom{]} & \phantom{[}\quad ] & [\quad \phantom{]} & \phantom{[}\quad ] &[\quad \phantom{]} &\phantom{[}\quad ] & [\quad \phantom{]} & \phantom{[}\quad ] & [\quad \phantom{]} & \phantom{[}\quad ]& [\quad \phantom{]} & \phantom{[}\quad ]& [\quad \phantom{]} &  $\cdots$   \\
 $\D_2$  & [\quad \phantom{]} & \phantom{[}\quad \phantom{]} & \phantom{[}\quad \phantom{]} & \phantom{[}\quad ] & [\quad \phantom{]} & \phantom{[}\quad \phantom{]} & \phantom{[}\quad \phantom{]} & \phantom{[}\quad ] &[\quad \phantom{]} &\phantom{[}\quad \phantom{]} & \phantom{[}\quad \phantom{]} & \phantom{[}\quad ] & [\quad \phantom{]} & \phantom{[}\quad \phantom{]}&\phantom{[}\quad \phantom{]} &\phantom{[}\quad ]  &[\quad \phantom{]} &  &  &$\cdots$   \\
  $\D_3$   & [\quad \phantom{]} & \phantom{[}\quad \phantom{]} & \phantom{[}\quad \phantom{]} & \phantom{[}\quad \phantom{]} &\phantom{[}\quad \phantom{]} &\phantom{[}\quad \phantom{]} & \phantom{[}\quad \phantom{]} & \phantom{[}\quad ] & [\quad \phantom{]} & \phantom{[}\quad \phantom{]}& & &   & &  & \phantom{[}\quad ] & [\quad \phantom{]} &  &  & $\cdots$   \\
\end{tabular}\vspace{-2ex}
\caption{Dense geometric covering (DGC) intervals.}
\label{fig:interval:dense}
\end{figure*}

Similar to existing adaptive algorithms, we will  run multiple instances of OGD over a set of intervals. Instead of using the GC intervals of  \citet{Adaptive:ICML:15}, we construct a dense version of GC intervals: 
\[
 \D= \bigcup_{k \in \N \cup \{0\} \& 2^k \leq T} \D_k
 \]
 where for all $k \in \N \cup \{0\}$, $\D_k=\left\{I_k^i = [ (i-1) \cdot 2^k+1, i \cdot 2^k]: i \in \N\right\}$. We present a graphical illustration of our dense geometric covering (DGC) intervals in Fig.~\ref{fig:interval:dense}. Compared with the original GC intervals, the main difference is that $\I_k$ in GC intervals is a partition of $\N \setminus \{1,\cdots, 2^k-1 \}$ to consecutive intervals of length $2^k$, while $\D_k$ in our DGC intervals is a partition of $\N$ to consecutive intervals of length $2^k$. Furthermore, we assume the total number of iterations $T$ is given beforehand, so we only construct intervals whose lengths are not larger than $T$.

For each interval $I_k^i \in \D$, we will run an instance of OGD. According to Theorem~\ref{thm:regret:OGD}, we set the step size as $\eta=D/(G \sqrt{2^k})$, which is able to minimize the static regret over $I_k^i$. For the initial solution, we choose one of the following two ways:
\begin{compactitem}
  \item If $i=1$, we set the initial solution as an arbitrary point in $\Omega$.
  \item If $i >1$, we set the initial solution as the last output of the OGD associated with $I_k^{i-1}$.
\end{compactitem}
In other words, the expert associated with each interval in $\D_k$, except the first one, is warm started by initiating OGD with the solution of the previous expert. 

To combine the actions of all experts, we choose the AdaNormalHedge \citep{pmlr-v40-Luo15} as our meta-algorithm. AdaNormalHedge is a parameter-free expert-tracking algorithm, which shares a similar regret bound as the sleeping CB  \citep{Improved:Strongly:Adaptive}, but with simpler updating rules. It makes use of a potential function:
\[
\Phi(R,C)=\exp\left(\frac{[R]_+^2}{3C}\right)
\]
where $[x]_+=\max(0,x)$ and $\Phi(0,0)$ is defined to be $1$, and a weight function with respect to this potential:
\[
w(R,C)= \frac{1}{2} \big(\Phi(R+1,C+1) - \Phi(R-1,C+1)\big).
\]

\begin{algorithm}[tb]
   \caption{Adaptive Online learning with Dynamic regret (AOD)}
   \label{alg:AOD}
\begin{algorithmic}[1]
   \FOR{$t=1$ {\bfseries to} $T$}
     \FOR{$I \in \C_t$}
        \STATE Create an expert $E_I$ which runs OGD, and set $R_{t-1,I} =  C_{t-1,I}=0$
        \STATE Set the step size of $E_I$ as $\eta=D/(G \sqrt{|I|})$
       \IF{$t=1$}
        \STATE Set the initial solution of $E_I$ arbitrarily
  \ELSE
  \STATE Identify the expert $E_J \in \A_t$ such that $|J|=|I|$
  \STATE Set the initial solution of $E_I$ to be the output of expert $E_J$, denoted by $\w_{t,J}$
  \STATE Remove $E_J$ from $\A_t$
  \ENDIF
  \STATE Add expert $E_I$ to the set of active experts $\A_t$
   \ENDFOR
   \STATE Receive the action $\w_{t,I}$ of each expert $E_I \in \A_t$, and calculate its weight $p_{t,I}$ according to (\ref{eqn:weight})
  \STATE Submit $\w_t$ defined in (\ref{eqn:wt}) and then receive  $f_t(\cdot)$
   \STATE For each $E_I \in \A_t$, update
   \[
   R_{t,I} = R_{t-1,I} + f_t(\w_t) - f_t(\w_{t,I}), \textrm{ and }    C_{t,I} = C_{t-1,I} + \left|f_t(\w_t) - f_t(\w_{t,I})\right|
   \]
   \STATE Pass $f_t(\cdot)$ to each expert $E_I \in \A_t$
   \ENDFOR
\end{algorithmic}
\end{algorithm}
The complete procedure is named as Adaptive Online learning with Dynamic regret (AOD), and summarized in Algorithm~\ref{alg:AOD}. We explain the main steps below. In each round $t$,  we will maintain a set of active experts, denoted by $\A_t$. To simplify the notation, let's define the set of intervals in $\D$ that start from $t$ as $\C_t$, i.e.,
\begin{equation} \label{eqn:c:t}
\C_t=\{I| I \in \D, t \in I, (t-1) \notin I\}.
\end{equation}
For each interval $I$ in $\C_t$, we  create an expert $E_I$ which runs an instance of OGD, and initialize two variables $R_{t-1,I}$ and $C_{t-1,I}$ that are used to calculate the weight of $E_I$ (Step 3). The step size of $E_I$ is set in Step 4, and the initial solution is set in Step 6 or 9. Note that when $t >1$, for each interval $I \in \C_t$, there must be an interval $E_J \in \A_t$ such that $|J|=|I|$ (Step 8).  Then, we use the output of $E_J$ to initialize $E_I$ (Step 9), and remove $E_J$ from $\A_t$ (Step 10). The new expert $E_I$ is added to $\A_t$ (Step 12).

In Step 14, we receive the action $\w_{t,I}$ of each active expert $E_I \in \A_t$, and assign the following weight to $E_I$
\begin{equation} \label{eqn:weight}
   p_{t,I}= \frac{w(R_{t-1,I},C_{t-1,I})}{\sum_{E_I \in \A_t} w(R_{t-1,I},C_{t-1,I}) }
\end{equation}
where
\[
\begin{split}
R_{t-1,I} = \sum_{u=\min I}^{t-1} f_u (\w_{u}) - f_u (\w_{u,I}),   \quad  C_{t-1,I}= &\sum_{u=\min I}^{t-1} \left|f_u (\w_{u}) - f_u (\w_{u,I})\right|,
\end{split}
\]
and $\min I$ denotes the starting round of interval $I$. In Step 15, we submit the weighted average
\begin{equation} \label{eqn:wt}
  \w_t = \sum_{E_I \in \A_t} p_{t,I}  \w_{t,I}
\end{equation}
as the output, and receive the loss function $f_t(\cdot)$. In Step 16, we update variables that are used to calculate probabilities in (\ref{eqn:weight}). Finally, we reveal the  function $f_t(\cdot)$ to all active experts so that they can make predictions for the next round.

We first present the strongly adaptive regret of  AOD.
\begin{thm} \label{thm:ada:AOD}
Under Assumptions~\ref{ass:1}, \ref{ass:2}, and \ref{ass:6}, the strongly adaptive regret of AOD  in Algorithm~\ref{alg:AOD} satisfies
\[
\begin{split}
\SAReg(T,\tau)\leq 8 \left(\sqrt{3 c(T)}+DG \right) \sqrt{\tau}=O\left(\sqrt{\tau \log T}\right)
\end{split}
\]
where
\begin{equation} \label{eqn:cT}
\begin{split}
c(T) \leq   1+ \ln T  + \ln(1+ \log_2 T) + \ln\frac{5+ 3 \ln(1+T)}{2}.
\end{split}
\end{equation}
\end{thm}
Note that our strongly adaptive regret matches the state-of-the-art result of \citet{Improved:Strongly:Adaptive} exactly. The main advantage is that AOD is also equipped with a dynamic regret bound, which is nearly optimal.

\begin{thm}\label{thm:dyn:AOD} Under Assumptions~\ref{ass:1}, \ref{ass:2}, and \ref{ass:6}, for \emph{any} comparator sequence $\u_1,\ldots,\u_T \in \Omega$,  AOD in Algorithm~\ref{alg:AOD} satisfies
\[
\begin{split}
\DReg(\u_1,\ldots,\u_T) \leq  &\left( \frac{3 D G}{2} + \frac{5G}{2} \sqrt{D P_T} +\sqrt{6c(T) \left(1+  \frac{2 P_T}{D} \right)} \right) \sqrt{T} 
\\
=& O\left( \sqrt{T (1+ P_T )\log T } \right)
\end{split}
\]
where $c(T)$ is given in (\ref{eqn:cT}).
\end{thm}
\textbf{Remark:} The dynamic regret  of AOD matches the $\Omega(\sqrt{T(1+P_T)})$  lower bound up to a logarithmic factor, and is slightly worse than the $O(\sqrt{T(1+P_T)})$ bound of Ader \citep{Adaptive:Dynamic:Regret:NIPS}. However, Ader is not equipped with any adaptive regret, while our AOD achieves the state-of-the-art adaptive regret as shown in Theorem~\ref{thm:ada:AOD}.

\subsection{The Second Method}
One limitation of AOD is that the total number of iterations $T$ needs to be known and fixed. In this section, we address this limitation by developing a three-layer algorithm, in which an additional layer is inserted to decouple the adaptive regret and the dynamic regret.

\begin{algorithm}[tb]
   \caption{Adaptive Online learning based on Ader (AOA)}
   \label{alg:AOA}
\begin{algorithmic}[1]
   \FOR{$t=1$ {\bfseries to} $T$}
     \FOR{$I \in \widetilde{\C}_t$}
        \STATE Create an expert $E_I$, which runs Ader, and set $R_{t-1,I} =  C_{t-1,I}=0$
     \STATE Pass the interval length $I$ to expert $E_I$
    \STATE Add expert $E_I$ to the set of active experts $\A_t$
   \ENDFOR
   \STATE Receive the action $\w_{t,I}$ of each expert $E_I \in \A_t$, and calculate its weight $p_{t,I}$ according to (\ref{eqn:weight})
  \STATE Submit $\w_t$ defined in (\ref{eqn:wt}) and  receive  $f_t(\cdot)$
     \STATE Remove experts whose ending times are $t$
    \[
    \A_{t} = \A_{t} \setminus \{ E_I | t \in I, (t+1) \notin I \}
    \]
   \STATE For each $E_I \in \A_t$, update
   \[
   \begin{split}
   R_{t,I} = R_{t-1,I} + f_t(\w_t) - f_t(\w_{t,I}), \textrm{ and }     C_{t,I} = C_{t-1,I} + \left|f_t(\w_t) - f_t(\w_{t,I})\right|
   \end{split}
   \]
   \STATE Pass $f_t(\cdot)$ to each expert $E_I \in \A_t$
   \ENDFOR
\end{algorithmic}
\end{algorithm}
\begin{algorithm}[tb]
   \caption{Adaptive learning for dynamic environment (Ader)}
   \label{alg:Ader}
   \begin{algorithmic}[1]
      \STATE {\bfseries Input:} The total number of iterations $T$
\STATE Construct the set $\H$ according to (\ref{eqn:def:H})
\STATE Create a set of experts $\{E_\eta| \eta \in \H \}$ by running OGD with each step size $\eta \in \H$
\STATE Sort step sizes in ascending order $\eta_1 \leq \eta_2 \leq \cdots \leq \eta_N$, and set $p_{1,\eta_i}=\frac{C}{i(i+1)}$ where  $C=1+\frac{1}{|\H|}$
\FOR{$t=1,\ldots,T$}
\STATE Receive the action $\w_{t,\eta}$ from each expert $E_\eta$
\STATE Submit
\[
\w_t = \sum_{\eta \in \H} p_{t,\eta} \w_{t, \eta}
\]
 and then receive  $f_t(\cdot)$
\STATE Update the weight of each expert by
\[
p_{t+1, \eta} = \frac{p_{t,\eta} e^{-\alpha f_t(\w_{t,\eta})} }{\sum_{\mu \in \H} p_{t,\mu} e^{-\alpha f_t(\w_{t,\mu})}}
\]
where $\alpha= \sqrt{8 /T}$
   \STATE Pass $f_t(\cdot)$ to each expert $E_\eta$
\ENDFOR
\end{algorithmic}
\end{algorithm}
The basic idea is very simple. Instead of running OGD as the expert in our previous AOD algorithm, we use Ader \citep{Adaptive:Dynamic:Regret:NIPS}, which is designed to minimize the dynamic regret, as the expert algorithm. The new algorithm is named as Adaptive Online learning based on Ader (AOA), and summarized in Algorithm~\ref{alg:AOA}. Because Ader itself is a two-layer algorithm, AOA is essentially a three-layer algorithm. The top layer takes responsibility for the adaptive regret, and the middle-layer is responsible for the dynamic regret. Because of this design, AOA is able to minimize the dynamic regret over any interval. In contrast,  the top layer in AOD takes care of both the adaptive regret and the dynamic regret.

We create experts based on the original GC intervals in (\ref{eqn:GC}) \citep{Adaptive:ICML:15}, because they can be constructed dynamically and do not need to know the total number of iterations $T$. Similar to the AOD algorithm, we use $\A_t$ to denote the set of active experts in round $t$, and $\widetilde{\C}_t$ to denote the set of intervals in $\I$ that start from $t$, i.e.,
\[
\widetilde{\C}_t=\{I| I \in \I, t \in I, (t-1) \notin I\}.
\]
For each interval $I$ in $\widetilde{\C}_t$, we will create an expert $E_I$ which runs an instance of Ader (Step 3),  pass the interval length $|I|$ to $E_I$ (Step 4), and add it to the set of active experts $\A_t$ (Step 5). As before, we combine the actions of all active experts by  AdaNormalHedge (Steps 7 and 8). After submitting  $\w_t$, AOA removes all the experts whose ending times are $t$ (Step 9). All the remaining steps of AOA are identical to those of AOD.

For the sake of completeness, we present the procedure of Ader in Algorithm~\ref{alg:Ader}, and give a brief introduction.  Ader takes the total number of iterations $T$ as the input, and constructs a set of step sizes
 \begin{equation} \label{eqn:def:H}
\H=\left\{ \left. \eta_i=\frac{2^{i-1} D}{G}\sqrt{\frac{7}{2T}} \right  | i=1,\ldots, N\right\}
\end{equation}
where $N= \lceil \frac{1}{2} \log_2 (1+4T/7) \rceil +1$ (Step 2). For each $\eta \in \H$, Ader creates an expert $E_\eta$ by running an instance of OGD with step size $\eta$ (Step 3).\footnote{The initial solution of OGD can be set arbitrarily.} The actions of  experts are combined by the standard Hedge algorithm (Steps 7 and 8) with nonuniform initial weights (Step 4) \citep{FREUND1997119,bianchi-2006-prediction}.

We present the theoretical guarantee of  AOA.
\begin{thm} \label{thm:AOA} Under Assumptions~\ref{ass:1}, \ref{ass:2}, and \ref{ass:6}, for \emph{any} interval $I=[r,s]\subseteq \N$ and \emph{any} comparator sequence $\u_r, \ldots, \u_s \in \Omega$,  AOA in Algorithm~\ref{alg:AOA} satisfies
\[
\begin{split}
\sum_{t=r}^s f_t(\w_t) - \sum_{t=r}^s f_t(\u_t)\leq & \left( 14\sqrt{ c'(s)} + 3  \left[1+ 2\ln (k_{I}+1)\right] + 23  DG \right) \sqrt{|I|}  + 5 G   \sqrt{ D P_{I} } \sqrt{|I| }\\
 = &O \left(\sqrt{|I| (\log s + P_{I})} \right)
\end{split}
\]
where
\[
\begin{split}
c'(s)  \leq & 1+ \ln s + \ln(1+ \log_2 s) + \ln\frac{5+ 3 \ln(1+s)}{2}, \\
P_I= & \sum_{t=r}^s \|\u_{t+1} - \u_{t}\|_2, \\
k_I = & \left\lfloor \frac{1}{2} \log_2 \left(1+ \frac{4 P_I}{7D} \right) \right \rfloor+1.
\end{split}
\]
\end{thm}
\textbf{Remark:} The above theorem indicates that our AOA algorithm can minimize the dynamic regret over any interval, which is a strong theoretical guarantee that allows us to derive either dynamic regret or adaptive regret. To derive dynamic regret over the whole interval, we set $I=[1,T]$, and then obtain an $O(\sqrt{T(\log T+P_T)})$ dynamic regret bound, which  nearly matches the $\Omega(\sqrt{T(1+P_T)})$ lower bound \citep{Adaptive:Dynamic:Regret:NIPS}, and becomes optimal when $P_T =\Omega( \log T)$. To derive adaptive regret, we set $\u_r =\cdots =\u_s$ such that $P_I=0$ and consider $s \leq T$, and then can prove $\SAReg(T,\tau) =O\left(\sqrt{\tau \log T}\right)$ which matches the state-of-the-art result \citep{Improved:Strongly:Adaptive} exactly.

\section{Analysis}
In this section, we present the analysis of our theoretical guarantees.
\subsection{Proof of Theorem~\ref{thm:rel}}
The proof of (\ref{eqn:dynamic:adap}) is similar to that of Theorem 3 of \citet{Dynamic:Regret:Adaptive}. We upper bound the dynamic regret in the following way
\begin{equation} \label{eqn:thm1:1}
\begin{split}
&\DReg(\u_1,\ldots,\u_T) \\
= &  \sum_{i=1}^k \left(\sum_{t=s_i}^{q_i} f_t(\w_t)  - \sum_{t=s_i}^{q_i} f_t(\u_t) \right)\\
= & \sum_{i=1}^k \left( \underbrace{\sum_{t=s_i}^{q_i} f_t(\w_t) -  \sum_{t=s_i}^{q_i} f_t(\v_i)}_{:=a_i} +\underbrace{\sum_{t=s_i}^{q_i} f_t(\v_i)- \sum_{t=s_i}^{q_i}  f_t(\u_t)}_{:=b_i} \right)
\end{split}
\end{equation}
where $\v_i$ could be any point in $\Omega$.

From the definition of strongly adaptive regret, we upper bound $a_i$ by
\[
\sum_{t=s_i}^{q_i} f_t(\w_t) -  \sum_{t=s_i}^{q_i} f_t(\v_i) \leq \SAReg(T,|I_i|)
\]
which is independent from the choice of $\v_i$. To bound $b_i$, we define the local path-length
\[
      P_T(i)=\sum_{t={s_i}}^{q_i-1} \|\u_{t+1} - \u_{t}\|_2.
\]
Setting $\v_i=\u_{s_i}$ and recalling that all the functions are $G$-lipschitz continuous, we have
\[
b_i  \leq  G \sum_{t=s_i}^{q_i} \|\u_{s_i}-\u_t\| \leq  G \sum_{t=s_i}^{q_i} P_T(i) = G |I_i| \cdot P_T(i).
\]
We obtain  (\ref{eqn:dynamic:adap}) by substituting the upper bounds of $a_i$ and $b_i$ into (\ref{eqn:thm1:1}).

We proceed to prove (\ref{eqn:dynamic:by:adap}), which is similar to the Corollary 5 of \citet{Dynamic:Regret:Adaptive}. As we shown in Theorem~\ref{thm:regret:OGD}, the static regret of online gradient descent  over any interval of
length $\tau$ is upper bounded by $DG \sqrt{\tau}$. Combining this fact with Theorem 2 of  \citet{Improved:Strongly:Adaptive},  we have the following theorem.
\begin{thm} \label{thm:coin:cb} Under Assumptions~\ref{ass:1} and \ref{ass:2}, the meta-algorithm of \citet{Improved:Strongly:Adaptive} satisifes
\[
\begin{split}
\SAReg(T,\tau) \leq \left(\frac{4DG}{\sqrt{2}-1} +8 \sqrt{7 \log T + 5} \right) \sqrt{\tau}  = O(\sqrt{\tau \log T}  )
 \end{split}
\]
for online convex optimization.
\end{thm}

To simplify the upper bound in (\ref{eqn:dynamic:adap}), we restrict to intervals of the same length $\tau$, and in this case $k=T/\tau$. Then, we have
\[
\begin{split}
\DReg(\u_1,\ldots,\u_T) \leq &  \min_{1 \leq \tau \leq T} \sum_{i=1}^k \big( \SAReg(T,\tau) + \tau G P_T(i) \big)\\
= & \min_{1 \leq \tau \leq T}  \left(  \frac{ \SAReg(T,\tau) T}{\tau} + \tau G \sum_{i=1}^k P_T(i) \right)\\
\leq & \min_{1 \leq \tau \leq T}  \left( \frac{ \SAReg(T,\tau) T}{\tau} + \tau G P_T \right).
\end{split}
\]
Combining with Theorem~\ref{thm:coin:cb}, we have
\[
\begin{split}
\DReg(\u_1,\ldots,\u_T) \leq  \min_{1 \leq \tau \leq T}  \left( \frac{(c +8 \sqrt{7 \log T})  T }{\sqrt{\tau}} + \tau G P_T \right)
\end{split}
\]
where
\[
c=\frac{4DG}{\sqrt{2}-1} +8\sqrt{5}.
\]

In the following, we consider two cases. If $P_T \geq D \sqrt{\log T/T}$, we choose
\[
\tau = \left( \frac{ D T \sqrt{\log T}}{P_T} \right)^{2/3} \leq T
\]
and have
\[
\DReg(\u_1,\ldots,\u_T)  \leq    \frac{(c +8 \sqrt{7 \log T} )  T^{2/3} P_T^{1/3} }{D^{1/3}\log^{1/6} T}    + G D^{2/3} T^{2/3} P_T^{1/3} \log^{1/3} T .
\]
Otherwise, we choose  $\tau=T$, and have
\[
\begin{split}
\DReg(\u_1,\ldots,\u_T)  \leq &(c +8 \sqrt{7 \log T})  \sqrt{T}  + G T P_T \\
  \leq & (c +8 \sqrt{7 \log T})  \sqrt{T}  + G D T  \sqrt{\frac{\log T}{T}} \\
  =& \left(c +8 \sqrt{7 \log T} + GD \sqrt{\log T} \right)   \sqrt{T}.
 \end{split}
\]

In summary, we have
\[
\begin{split}
\DReg(\u_1,\ldots,\u_T)  \leq & \max\left\{\begin{split}
&  \left(c +8 \sqrt{7 \log T} + GD \sqrt{\log T} \right)   \sqrt{T}  \\
&  \frac{(c +8 \sqrt{7 \log T} )  T^{2/3} P_T^{1/3} }{D^{1/3}\log^{1/6} T}    + G D^{2/3} T^{2/3} P_T^{1/3} \log^{1/3} T
\end{split} \right.\\
= & O \left( \max\left\{  \sqrt{T \log T} , T^{2/3} P_T^{1/3} \log^{1/3} T  \right\} \right).
\end{split}
\]
\subsection{Proof of Theorem~\ref{thm:ada:AOD}}
We first present the meta-regret of AOD.  Let $m(t)$ be the total number of experts created up to round $t$. It is easy to verify that
\[
m(t) \leq t (1 + \log_2 T ).
\]
Then, according to Theorems~1\&3 of \citet{pmlr-v40-Luo15} and Jensen's inequality \citep{Convex-Optimization}, we have the following lemma.
\begin{lemma} \label{lemma:AOD} Under Assumption~\ref{ass:6}, for any interval $J=[i,j] \in \D$, AOD satisfies
\[
\sum_{u=i}^t f_u(\w_u) - \sum_{u=i}^t f_u(\w_{u,J})  \leq \sqrt{3 (t-i+1) c(t) }, \ \forall t \in J
\]
where
\[
\begin{split}
c(t) \leq &1+ \ln m(t) + \ln\frac{5+ 3 \ln(1+t)}{2}\\
 \leq & 1+ \ln t + \ln(1+ \log_2 T) + \ln\frac{5+ 3 \ln(1+t)}{2} .
\end{split}
\]
\end{lemma}

We proceed to bound the adaptive regret of AOD. To this end, we first bound the regret of AOD over any interval $J=[i,j] \in \D$. By combining the meta-regret in Lemma~\ref{lemma:AOD} and the expert-regret in Theorem~\ref{thm:regret:OGD}, we immediately have the following bound.
\begin{lemma} \label{lem:AOD:D} Under Assumptions~\ref{ass:1}, \ref{ass:2}, and \ref{ass:6}, for any interval $J=[i,j] \in \D$, AOD satisfies
\[
\sum_{t \in J} f_t(\w_t) -  \min_{\w \in \Omega} \sum_{t \in J} f_t(\w) \leq  \left(\sqrt{3 c(j)}+DG \right) \sqrt{|J|}.
\]
\end{lemma}
We then extend the above regret bound to any interval $I=[r,s] \subseteq [T]$. To this end, we need the following lemma about the DGC intervals, which has a similar property as the original GC intervals \citep{Adaptive:ICML:15}.

\begin{lemma} \label{lem:DGC:intervals} For any interval $[r,s] \subseteq [T]$, it can be partitioned into two sequences of disjoint and consecutive intervals, denoted by $I_{-p},\ldots,I_0 \in \D$ and $I_1,\ldots,I_q \in \D$, such that
\[
|I_{-i}|/ |I_{-i+1}| \leq 1/2, \ \forall i \geq 1
\]
and
\[
|I_i|/|I_{i-1}| \leq 1/2, \ \forall i \geq 2.
\]
\end{lemma}
Then, based on Lemmas~\ref{lem:AOD:D} and~\ref{lem:DGC:intervals}, we bound the regret with respect to any $\w \in \Omega$ over $I=[r,s]$ in the following way
\[
\begin{split}
&\sum_{t=r}^s f_t(\w_t) - \sum_{t=r}^s f_t(\w) = \sum_{i=-p}^{q} \left(\sum_{t \in I_i} f_t(\w_t) - \sum_{t \in I_i} f_t(\w)  \right) \\
\leq & \sum_{i=-p}^{q} \left(\sqrt{3 c(s)}+DG \right) \sqrt{|I_i|}\leq 2 \left(\sqrt{3 c(s)}+DG \right) \sum_{i=0}^\infty (2^{-i} |I|)^{1/2} 
\leq  8 \left(\sqrt{3 c(s)}+DG \right)  \sqrt{I}.
\end{split}
\]
Thus, the strongly adaptive regret
\[
\begin{split}
\SAReg(T,\tau) = \max_{[s, s+\tau -1] \subseteq [T]} \left(\sum_{t=s}^{s+\tau -1}  f_t(\w_t)  - \min_{\w \in \Omega}  \sum_{t=s}^{s+\tau -1} f_t(\w) \right) \leq  8 \left(\sqrt{3 c(T)}+DG \right) \sqrt{\tau}.
\end{split}
\]
\subsection{Proof of Lemma~\ref{lem:DGC:intervals}}
\begin{figure*}
\centering
\begin{tabular}{@{}c@{\hspace{1ex}}*{19}{@{\hspace{0.6ex}}c}@{\hspace{1ex}}c@{}}
$t$ & 1 & 2 & 3 & 4 & 5 & 6 & 7 & 8 & 9 & 10 &11 &12 & 13 & 14 &15 & 16 & 17 & 18 &19  &$\cdots$ \\
 $\I_0$ & [\quad ] & [\quad ] &  [\quad ] & [\quad ] & [\quad ] & [\quad ] & [\quad ] & [\quad ] & [\quad ] & [\quad ] & [\quad ] &[\quad ] &[\quad ] & [\quad ] & [\quad ] & [\quad ]& [\quad ]&[\quad ] & [\quad ]&  $\cdots$   \\
 $\I_1$ &  & [\quad  \phantom{]}& \phantom{[} \quad ] & [\quad \phantom{]} & \phantom{[}\quad ] & [\quad \phantom{]} & \phantom{[}\quad ] & [\quad \phantom{]} & \phantom{[}\quad ] & [\quad \phantom{]} & \phantom{[}\quad ] &[\quad \phantom{]} &\phantom{[}\quad ] & [\quad \phantom{]} & \phantom{[}\quad ] & [\quad \phantom{]} & \phantom{[}\quad ]& [\quad \phantom{]} & \phantom{[}\quad ]&  $\cdots$   \\
 $\I_2$  &  & &  & [\quad \phantom{]} & \phantom{[}\quad \phantom{]} & \phantom{[}\quad \phantom{]} & \phantom{[}\quad ] & [\quad \phantom{]} & \phantom{[}\quad \phantom{]} & \phantom{[}\quad \phantom{]} & \phantom{[}\quad ] &[\quad \phantom{]} &\phantom{[}\quad \phantom{]} & \phantom{[}\quad \phantom{]} & \phantom{[}\quad ] & [\quad \phantom{]} & \phantom{[}\quad \phantom{]}&\phantom{[}\quad \phantom{]} &\phantom{[}\quad ] &$\cdots$   \\
  $\I_3$   &  & &  &  &  &  &  & [\quad \phantom{]} & \phantom{[}\quad \phantom{]} & \phantom{[}\quad \phantom{]} & \phantom{[}\quad \phantom{]} &\phantom{[}\quad \phantom{]} &\phantom{[}\quad \phantom{]} & \phantom{[}\quad \phantom{]} & \phantom{[}\quad ] & [\quad \phantom{]} & \phantom{[}\quad \phantom{]}& & & $\cdots$   \\
\end{tabular}\vspace{-2ex}
\caption{Geometric covering (GC) intervals of \citet{Adaptive:ICML:15}.}
\label{fig:interval:saol}
\end{figure*}

In the analysis, we will make use of the covering property of GC intervals. To this end, we provide a graphical illustration of GC intervals in Fig.~\ref{fig:interval:saol}.

We first consider any interval $[r,s]$ where $r \geq 2$. Let $\widehat{\D}$ be the subset of $\D$ where all intervals that start from $1$ are removed, i.e.,
\[
\widehat{\D}=\{ I: I \in \D, \ 1 \notin  I\}.
\]
Then, covering $[r,s]$ by $\D$ is equivalent to covering $[r,s]$ by $\widehat{\D}$, since $r \geq 2$.  Comparing Fig.~\ref{fig:interval:saol} and Fig.~\ref{fig:interval:dense}, it is easy to see that the structure of $\widehat{\D}$ is identical to that of $\C$. So, we can directly re-use the covering property of GC intervals. Specifically, from Lemma 1.2 of \citet{Adaptive:ICML:15}, we  have the following lemma.
\begin{lemma} \label{lem:cover:special} For any interval $[r,s] \subseteq [T]$ where $r \geq 2$, it can be partitioned into two sequences of disjoint and consecutive intervals, denoted by $I_{-p},\ldots,I_0 \in \D$ and $I_1,\ldots,I_q \in \D$, such that
\begin{equation} \label{eqn:com:1}
|I_{-i}|/ |I_{-i+1}| \leq 1/2, \ \forall i \geq 1
\end{equation}
and
\begin{equation} \label{eqn:com:2}
|I_i|/|I_{i-1}| \leq 1/2, \ \forall i \geq 2.
\end{equation}
\end{lemma}

Next, we consider any interval $[1,s]$. To this end, we first apply Lemma~\ref{lem:cover:special} to interval $[2,s]$. As a result, we find two sequences of disjoint and consecutive intervals, denoted by $I_{-p},\ldots,I_0 \in \D$ and $I_1,\ldots,I_q \in \D$, such that (\ref{eqn:com:1}) and (\ref{eqn:com:2}) hold. Then, $[1,s]$ can be covered in the following way.

We construct a new interval $I_{-p-1}=[1,1]$. If the new sequence $I_{-p-1}, I_{-p},\ldots,I_0 \in \D$ satisfies (\ref{eqn:com:1}), we have finished, and return $I_{-p-1}, I_{-p},\ldots,I_0 \in \D$  and $I_1,\ldots,I_q \in \D$. Otherwise, i.e., $|I_{-p-1}|=|I_{-p}|$, we keep on merging the first two intervals in $I_{-p-1}, I_{-p},\ldots,I_0$  until (\ref{eqn:com:1}) is satisfied.

\subsection{Proof of Theorem~\ref{thm:dyn:AOD}}
We establish the dynamic regret bound by showing that for any possible value of the path-length $P_T$, there exists an interval set $\D_k=\{I_k^1,I_k^2,\ldots\} \subseteq \D$ such that
\begin{compactenum}[(i)]
  \item the combination of experts $E_{I_k^1},E_{I_k^2},\ldots$ enjoys a tight dynamic regret with respect to the comparator sequence $\u_1,\ldots,\u_T$.
  \item the meta-regret of AOD with respect to the expert sequence $E_{I_k^1},E_{I_k^2},\ldots$ is also well-bounded.
\end{compactenum}

Note that
\[
 P_T=\sum_{t=1}^T \|\u_{t+1} - \u_{t}\|_2  \in [0, DT].
\]
We will study $P_T \in (D, DT]$ and $P_T \in [0,D]$ separately. For the first case, we construct $s = \lceil\log_2 T\rceil$ consecutive intervals
\[
\delta_i=\left(D 2^{i-1},  D 2^i\right] , \ i =1,\ldots,s
\]
such that
\[
(D, DT] \subseteq \bigcup_{i=1}^s \delta_i .
\]
Then, we investigate the dynamic regret when $P_T$ belongs to each $\delta_i$.  Note that
\[
\D=  \D_0 \cup \D_1 \cdots \D_{\lfloor\log_2 T\rfloor}.
\]
Thus, for each $\delta_i$, there is a corresponding interval set $\D_{s-i} \subseteq \D$.  We  make use of $\D_{s-i}$ to bound the dynamic regret, and prove the following lemma.
\begin{lemma} \label{lem:bound:deltai} Suppose $P_T \in \delta_i$, and $i \in [s]$. Then, we have
\begin{equation} \label{eqn:bound:first}
\begin{split}
\sum_{t=1}^T f_t(\w_{t}) - \sum_{t=1}^T  f_t(\u_t) \leq  \left( \frac{D G}{2} + \frac{3G}{2} \sqrt{2 D P_T} +\sqrt{3 c(T) \left(1+  \frac{2 P_T}{D} \right)} \right) \sqrt{T}.
\end{split}
\end{equation}
\end{lemma}

Finally, we consider the case $P_T \in [0, D]$. We  make use of the interval set $\D_{\lfloor\log_2 T\rfloor}$ to bound the dynamic regret, and establish the following lemma.
 \begin{lemma} \label{lem:bound:last} Suppose $P_T \leq D$. Then, we have
\begin{equation} \label{eqn:bound:second}
\begin{split}
\sum_{t=1}^T f_t(\w_{t}) - \sum_{t=1}^T  f_t(\u_t) \leq  \left(\frac{1+\sqrt{2}}{2} DG +  G \sqrt{ D P_T} +\sqrt{6 c(T) }   \right) \sqrt{T}.
\end{split}
\end{equation}
\end{lemma}

We complete the proof by combining (\ref{eqn:bound:first}) and (\ref{eqn:bound:second}).

\subsection{Proof of Lemma~\ref{lem:bound:deltai}}
We first introduce the dynamic regret of OGD \citep{zinkevich-2003-online}. Specifically, the following result can be distilled from  \citet{Adaptive:Dynamic:Regret:NIPS}.
\begin{thm} \label{thm:OGD:dynamic} Under Assumptions~\ref{ass:1} and \ref{ass:2}, we have
\[
\sum_{t=1}^T f_t(\w_{t}) - \sum_{t=1}^T  f_t(\u_t) \leq \frac{1}{2 \eta}\left( \|\w_1-\u_1\|_2^2 -\|\w_{T+1}-\u_{T+1}\|_2^2\right) + \frac{D}{\eta} \sum_{t=1}^T \|\u_{t+1}-\u_t\|_2 + \frac{\eta T}{2 } G^2
\]
for any comparator sequence $\u_1,\ldots,\u_T \in \Omega$.
\end{thm}

Recall that the length of each interval in $\D_{s-i}$ is $2^{s-i}$. Then, the interval $[1,T]$ can be covered by intervals
\[
I_{s-i}^1,\ldots,I_{s-i}^{m}
\]
where $m= \lceil T /2^{s-i} \rceil$. To simplify the notation, we denote the starting time and ending time of $I_{s-i}^{u}$ by
\[
s^u=(u-1) \cdot 2^{s-i}+1, \textrm{ and } e^u=  u \cdot 2^{s-i}.
\]

Then, we decompose the dynamic regret of AOD as
\begin{equation}\label{eqn:dyn:decompose}
\begin{split}
&\sum_{t=1}^T f_t(\w_{t}) - \sum_{t=1}^T  f_t(\u_t) \\
=& \sum_{u=1}^{m-1} \left(\sum_{t=s^u}^{e^u}  \Big( f_t(\w_{t}) -   f_t(\u_t) \Big)\right) + \sum_{t= s^m}^T \Big( f_t(\w_{t}) -   f_t(\u_t) \Big)\\
=&  \underbrace{\sum_{u=1}^{m-1} \left(\sum_{t=s^u}^{e^u}  \Big( f_t(\w_{t}) -  f_t(\w_{t,I_{s-i}^u}) \Big)\right) + \sum_{t= s^m}^T \Big( f_t(\w_{t}) -   f_t(\w_{t, I_{s-i}^{m}}) \Big)}_{B}\\
& +  \underbrace{\sum_{u=1}^{m-1} \left(\sum_{t=s^u}^{e^u}  \Big( f_t(\w_{t,I_{s-i}^u}) -   f_t(\u_t) \Big)\right) + \sum_{t= s^m}^T \Big( f_t(\w_{t, I_{s-i}^{m}}) -   f_t(\u_t) \Big)}_{A}.
\end{split}
\end{equation}
We proceed to bound the $A$, which is the dynamic regret of the experts with respect to $\u_1,\ldots,\u_T$. Note that all the experts run an instance of OGD with the same step size $\eta$.  According to Theorem~\ref{thm:OGD:dynamic}, we have
\begin{equation}\label{eqn:dyn:1}
\begin{split}
&A \\
\leq &\sum_{u=1}^{m-1} \left( \frac{\|\w_{s^u,I_{s-i}^u}-\u_{s^u}\|_2^2 -\|\w_{e^u+1,I_{s-i}^u}-\u_{e^u+1}\|_2^2}{2 \eta} + \frac{D}{\eta} \sum_{t=s^u}^{e^u} \|\u_{t+1}-\u_t\|_2 + \frac{\eta (e^u-s^u+1) G^2}{2 }  \right)  \\
&+  \left( \frac{\|\w_{s^m,I_{s-i}^m}-\u_{s^m}\|_2^2 -\|\w_{T+1,I_{s-i}^m}-\u_{T+1}\|_2^2}{2 \eta} + \frac{D}{\eta} \sum_{t=s^m}^T \|\u_{t+1}-\u_t\|_2 + \frac{\eta (T-s^m+1)G^2}{2 } \right)\\
=&  \frac{1}{2 \eta}\left( \|\w_{1,I_{s-i}^1}-\u_1\|_2^2 -\|\w_{T+1,I_{s-i}^m}-\u_{T+1}\|_2^2\right) + \frac{D}{\eta} \sum_{t=1}^T \|\u_{t+1}-\u_t\|_2 + \frac{\eta T G^2}{2 }  \\
\leq & \frac{D^2}{2 \eta}  + \frac{D}{\eta} \sum_{t=1}^T \|\u_{t+1}-\u_t\|_2 + \frac{\eta T}{2 } G^2 \\
= &\frac{D G}{2} \sqrt{2^{s-i}} + G P_T \sqrt{2^{s-i}}+ \frac{DG T}{2 \sqrt{2^{s-i}}}
\end{split}
\end{equation}
where the first equality follows from the fact  $e^u+1=s^{u+1}$ and $\w_{e^u+1,I_{s-i}^u}=\w_{s^{u+1},I_{s-i}^{u+1}}$ (warm start), and the second equality is due to  $\eta=D/(G \sqrt{2^{s-i}})$.  From the fact $D 2^{i-1} < P_T \leq D 2^i$ and $s = \lceil\log_2 T\rceil$, we have
\begin{equation} \label{eqn:dyn:2}
2^{s-i} \leq T, \ \sqrt{2^{s-i}} \leq \sqrt{2 T \frac{D}{P_T}}  \textrm{ and } \sqrt{2^{s-i}} \geq \sqrt{ T \frac{D}{2 P_T}}.
\end{equation}
Combining (\ref{eqn:dyn:1}) with (\ref{eqn:dyn:2}), we have
\begin{equation} \label{eqn:bound:A}
A \leq \frac{D G}{2} \sqrt{ T }  + G P_T \sqrt{2 T \frac{D}{P_T}} + \frac{DG T}{2 } \sqrt{ \frac{2 P_T}{T D}}= \left(\frac{D G}{2} + \frac{3G}{2} \sqrt{2 D P_T}  \right) \sqrt{T}.
\end{equation}

Next, we bound $B$, which is the meta-regret of AOD with respect to experts. According to Lemma~\ref{lemma:AOD}, we have
\begin{equation} \label{eqn:bound:B}
\begin{split}
B \leq & \sum_{u=1}^{m-1} \left(   \sqrt{3 (e^u-s^u+1) c(e^u) } \right) + \sum_{t= s^m}^T \sqrt{3 (T-s^m+1) c(T) } \\
\leq & \sqrt{3 c(T) } \left( \sum_{u=1}^{m-1} \sqrt{(e^u-s^u+1)} + \sqrt{(T-s^m+1)} \right) \leq \sqrt{3 c(T) m T } \\
\leq & \sqrt{3 c(T) \left(1+  \frac{2 P_T}{D} \right)} \sqrt{T}
\end{split}
\end{equation}
where the last step is due to
\[
m \leq 1+  \frac{T}{2^{s-i}} \overset{\text{(\ref{eqn:dyn:2})}}{\leq}  1+  \frac{2 P_T}{D}.
\]
We complete the proof by combining (\ref{eqn:dyn:decompose}), (\ref{eqn:bound:A})  and (\ref{eqn:bound:B}).

\subsection{Proof of Lemma~\ref{lem:bound:last}}
The proof is similar to that of Lemma~\ref{lem:bound:deltai}. To simplify the notation, we define $\alpha=\lfloor\log_2 T\rfloor$. Following (\ref{eqn:dyn:decompose}), we decompose the dynamic regret as
\begin{equation} \label{eqn::bound:last:1}
\begin{split}
&\sum_{t=1}^T f_t(\w_{t}) - \sum_{t=1}^T  f_t(\u_t) \\
=& \underbrace{\sum_{t=1}^{2^\alpha}  \Big( f_t(\w_{t}) -  f_t(\w_{t,I_{\alpha}^1}) \Big) + \sum_{t= 2^\alpha+1}^T \Big( f_t(\w_{t}) -   f_t(\w_{t, I_{\alpha}^{2}}) \Big)}_{B}\\
& + \underbrace{\sum_{t=1}^{2^\alpha}   \Big( f_t(\w_{t,I_{\alpha}^1}) -   f_t(\u_t) \Big) + \sum_{t= 2^\alpha+1}^T  \Big( f_t(\w_{t, I_{\alpha}^{2}}) -   f_t(\u_t) \Big)}_{A}.
\end{split}
\end{equation}
Following the derivation of (\ref{eqn:dyn:1}), we bound $A$ as
\begin{equation} \label{eqn::bound:last:2}
\begin{split}
A \leq &\frac{D G}{2} \sqrt{2^{\alpha}} + G P_T \sqrt{2^{\alpha}}+ \frac{DG T}{2 \sqrt{2^{\alpha}}} \\
 \leq &\left( \frac{1+\sqrt{2}}{2} DG +  G P_T \right) \sqrt{T} \leq \left( \frac{1+\sqrt{2}}{2} DG +  G \sqrt{ D P_T } \right) \sqrt{T}
\end{split}
\end{equation}
where we use the fact $\frac{T}{2} \leq 2^\alpha \leq T$. Following the derivation of (\ref{eqn:bound:B}), we upper bound $B$ as
\begin{equation} \label{eqn::bound:last:3}
B \leq \sqrt{6 c(T) T }.
\end{equation}
We complete the proof by substituting  (\ref{eqn::bound:last:2}) and (\ref{eqn::bound:last:3}) into (\ref{eqn::bound:last:1}).

\subsection{Proof of Theorem~\ref{thm:AOA}}
We present two theoretical guarantees that support our analysis. The first is the dynamic regret of Ader.
\begin{thm}[Theorem 3 of \citet{Adaptive:Dynamic:Regret:NIPS}] \label{thm:ader:dynamic} Under Assumptions~\ref{ass:1}, \ref{ass:2}, and \ref{ass:6}, for \emph{any} comparator sequence $\u_1,\ldots,\u_T \in \Omega$,   Ader  satisfies
\[
\begin{split}
\sum_{t=1}^T f_t(\w_t) -  \sum_{t=1}^T f_t(\u_t) \leq \frac{3G}{4} \sqrt{2 T (7D^2 + 4 DP_T) }+ \frac{\sqrt{2T} }{4} \left[1+ 2\ln (k+1)\right] \\
\end{split}
\]
where
\begin{equation} \label{eqn:def:k}
k = \left\lfloor \frac{1}{2} \log_2 \left(1+ \frac{4P_T}{7D} \right) \right \rfloor+1.
\end{equation}
\end{thm}
The second is the property of the GC intervals.
\begin{lemma}[Lemma 1.2 of \citet{Adaptive:ICML:15}]  \label{lem:GC:intervals} For any interval $[r,s] \subseteq \N$, it can be partitioned into two sequences of disjoint and consecutive intervals, denoted by $I_{-p},\ldots,I_0 \in \I$ and $I_1,\ldots,I_q \in \I$, such that
\[
|I_{-i}|/ |I_{-i+1}| \leq 1/2, \ \forall i \geq 1
\]
and
\[
|I_i|/|I_{i-1}| \leq 1/2, \ \forall i \geq 2.
\]
\end{lemma}

Similar to the proof of Theorem~\ref{thm:ada:AOD}, we first bound the meta-regret of AOA.  Let $m'(t)$ be the total number of experts created up to round $t$. Then, we have
\[
m'(t) \leq t (1 + \log_2 t )
\]
because the active expert in the $t$-round is smaller than $1+\log_2 t$ \citep{Adaptive:ICML:15}.  From Theorem 1 of \citet{pmlr-v40-Luo15} and Jensen's inequality \citep{Convex-Optimization}, we have the following lemma.
\begin{lemma} \label{lemma:AOA} Under Assumption~\ref{ass:6}, for any interval $J=[i,j] \in \I$, AOA satisfies
\[
\sum_{t=i}^j f_t(\w_t) - \sum_{t=i}^j f_t(\w_{t,J})  \leq \sqrt{3 |J| c'(j) }
\]
where
\[
c'(j) \leq 1+ \ln m'(j) + \ln\frac{5+ 3 \ln(1+j)}{2} \leq 1+ \ln j + \ln(1+ \log_2 j) + \ln\frac{5+ 3 \ln(1+j)}{2} .
\]
\end{lemma}
Combining Lemma~\ref{lemma:AOA} with Theorem~\ref{thm:ader:dynamic}, we can bound the dynamic regret of AOA over any interval $J=[i,j] \in \I$.
\begin{lemma} \label{lem:dynamic:regret} Under Assumptions~\ref{ass:1}, \ref{ass:2}, and \ref{ass:6}, for \emph{any} interval $J=[i,j] \in \I$  and  \emph{any} comparator sequence $\u_i,\ldots,\u_j \in \Omega$, AOA satisfies
\[
\sum_{t=i}^j f_t(\w_t) - \sum_{t=i}^j f_t(\u_t)  \leq \sqrt{3 |J| c'(j)} + \frac{3G}{4} \sqrt{2 |J| (7D^2 + 4 D P_J) }+ \frac{\sqrt{2|J|} }{4} \left[1+ 2\ln (k_J+1)\right] \\
\]
where
\[
P_J=\sum_{t=i}^j \|\u_{t+1} - \u_{t}\|_2, \textrm{ and } k_J = \left\lfloor \frac{1}{2} \log_2 \left(1+ \frac{4 P_J}{7D} \right) \right \rfloor+1.
\]
\end{lemma}

Next, we extend the above dynamic regret bound to any interval $I=[r,s] \subseteq \N$ by utilizing Lemma~\ref{lem:GC:intervals}. We first decompose the dynamic regret over $I=[r,s]$ as
\[
\begin{split}
&\sum_{t=r}^s f_t(\w_t) - \sum_{t=r}^s f_t(\u_t)\\
 =& \underbrace{\sum_{i=-p}^{0} \left(\sum_{t \in I_i} f_t(\w_t) - \sum_{t \in I_i} f_t(\u_t)  \right)}_{A}+ \underbrace{\sum_{i=1}^{q} \left(\sum_{t \in I_i} f_t(\w_t) - \sum_{t \in I_i} f_t(\u_t)  \right)}_{B} .
\end{split}
\]
We proceed to bound $A$ based on Lemma~\ref{lem:dynamic:regret}, and have
\[
\begin{split}
&A \\
\leq  & \sum_{i=-p}^{0} \left(\sqrt{3 |I_i| c'(s)} + \frac{3G}{4} \sqrt{2 |I_i| (7D^2 + 4 D P_{I_i}) }+ \frac{\sqrt{2|I_i|} }{4} \left[1+ 2\ln (k_{I}+1)\right] \right)  \\
\leq & \left( \sqrt{3  c'(s)} + \frac{\sqrt{2} }{4} \left[1+ 2\ln (k_{I}+1)\right] + \frac{3 \sqrt{14} DG}{4} \right) \sum_{i=-p}^{0} \sqrt{|I_i|} + \frac{3 \sqrt{2D} G}{2}   \sum_{i=-p}^{0} \sqrt{P_{I_i} |I_i|  } \\
\leq & \left( \sqrt{3  c'(s)} + \frac{\sqrt{2} }{4} \left[1+ 2\ln (k_{I}+1)\right] + \frac{3 \sqrt{14} DG}{4} \right) \sum_{i=0}^\infty (2^{-i} |I|)^{1/2}  + \frac{3 \sqrt{2D} G}{2}    \sqrt{ \sum_{i=-p}^{0} P_{I_i} }  \sqrt{ \sum_{i=-p}^{0}|I_i| }\\
\leq & \left( 4\sqrt{3  c'(s)} + \sqrt{2}  \left[1+ 2\ln (k_{I}+1)\right] + 3 \sqrt{14} DG \right) \sqrt{|I|}  + \frac{3 \sqrt{2D} G}{2}   \sqrt{  P_{I} } \sqrt{|I| } .
\end{split}
\]
Notice that $B$ can be bounded in the same way, and thus
\[
\begin{split}
&\sum_{t=r}^s f_t(\w_t) - \sum_{t=r}^s f_t(\u_t)\\
\leq&  \left( 8\sqrt{3  c'(s)} + 2\sqrt{2}  \left[1+ 2\ln (k_{I}+1)\right] + 6 \sqrt{14} DG \right) \sqrt{|I|}  + 3 \sqrt{2} G   \sqrt{ D P_{I} } \sqrt{|I| } \\
\leq&  \left( 14\sqrt{ c'(s)} + 3  \left[1+ 2\ln (k_{I}+1)\right] + 23  DG \right) \sqrt{|I|}  + 5 G   \sqrt{ D P_{I} } \sqrt{|I| } .
\end{split}
\]

\section{Conclusion and Future Work}
Inspired by recent developments of dynamic regret and adaptive regret, this paper asks whether it is possible to bound them simultaneously. We provide affirmative answers by proposing novel algorithms that achieve this goal. The first method, namely AOD, runs multiple instances of OGD over specifically designed intervals, uses warm start to connect successive OGD's, and then combines multiple decisions by an expert-tracking algorithm. Theoretical analysis shows that AOD enjoys a tight adaptive regret and a nearly optimal dynamic regret. The second method, namely AOA, maintains multiple instances of Ader, and combines them in the same way as AOD. We demonstrate that AOA is equipped with a strong theoretical guarantee in the sense that it can minimize the dynamic regret over any interval.

One way to extend our work is to use the curvature of functions, such as smoothness and strong convexity \citep{ML:Hazan:2007,NIPS2010_Smooth}, to further tighten our upper bounds. The main challenge is that improving the dynamic regret for all comparator sequences is very difficult, and will be investigated in the future. Another future work is to establish data-dependent bounds \citep{JMLR:Adaptive} for dynamic regret and adaptive regret  in the hope that the  structure of data, such as sparseness, can be exploited to improve the performance.

\bibliography{E:/MyPaper/ref}
\end{document}